\documentclass[10pt,twocolumn,letterpaper]{article}
\usepackage[pagenumbers]{cvpr} 
\def\method{SPAR3D\xspace}
\def\gentime{0.7\xspace}
\def\point{\boldsymbol{p}}
\def\condition{\boldsymbol{c}}
\definecolor{cvprblue}{rgb}{0.21,0.49,0.74}
\usepackage[pagebackref,breaklinks,colorlinks,allcolors=cvprblue]{hyperref}
\usepackage{comment}

\makeatletter
\renewcommand{\paragraph}{%
  \@startsection{paragraph}{4}%
  {\z@}{1ex \@plus 1ex \@minus .2ex}{-1em}%
  {\normalfont\normalsize\bfseries}%
}
\makeatother

\title{\method: Stable Point-Aware Reconstruction of 3D Objects from Single Images}

\author{Zixuan Huang$^{1,2*}$ \quad Mark Boss$^{1}$ \quad Aaryaman Vasishta$^{1}$ \quad James M. Rehg$^{2}$ \quad Varun Jampani$^{1}$\\
$^1$Stability AI, $^2$UIUC
}

\begin{document}
\twocolumn[{%
\renewcommand\twocolumn[1][]{#1}%
\maketitle
\begin{center}
    \centering
    \captionsetup{type=figure}
    \includegraphics[width=1.0\textwidth]{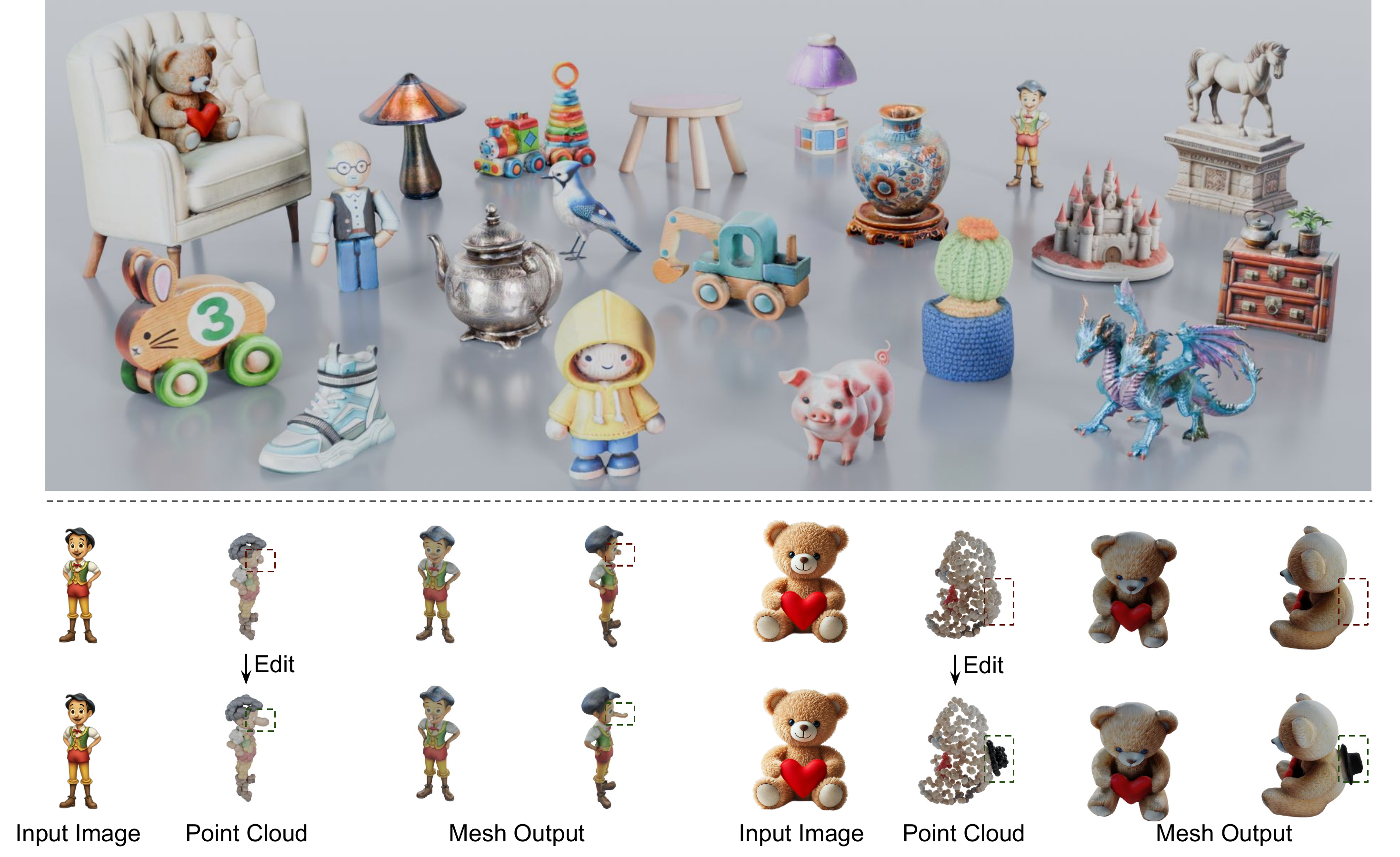}
    \captionof{figure}{We present \method, a state-of-the-art 3D reconstructor that reconstructs high-quality 3D meshes from single-view images. \method enjoys a fast reconstruction speed at \gentime seconds and supports interactive user edits.
    }
    \vspace{3mm}
\label{fig:teaser}
\end{center}%
}]

\iftoggle{cvprfinal}{
\def\thefootnote{*}\footnotetext{Work done at Stability AI.}\def\thefootnote{\arabic{footnote}}
}

\begin{abstract}
We study the problem of single-image 3D object reconstruction. Recent works have diverged into two directions: regression-based modeling and generative modeling. Regression methods efficiently infer visible surfaces, but struggle with occluded regions. Generative methods handle uncertain regions better by modeling distributions, but are computationally expensive and the generation is often misaligned with visible surfaces. In this paper, we present \method, a novel two-stage approach aiming to take the best of both directions. The first stage of \method generates sparse 3D point clouds using a lightweight point diffusion model, which has a fast sampling speed. The second stage uses both the sampled point cloud and the input image to create highly detailed meshes. Our two-stage design enables probabilistic modeling of the ill-posed single-image 3D task while maintaining high computational efficiency and great output fidelity. Using point clouds as an intermediate representation further allows for interactive user edits. Evaluated on diverse datasets, \method demonstrates superior performance over previous state-of-the-art methods, at an inference speed of \gentime seconds.
Project page with code and model: \url{https://spar3d.github.io}
\end{abstract}    
\section{Introduction}
\label{sec:intro}

Reconstructing 3D objects from monocular images is a fundamental problem in computer vision. An efficient reconstruction system opens up a wide range of applications, including augmented reality, filmmaking, and manufacturing. Monocular 3D reconstruction is also a complex inverse problem: while the visible surface can be estimated from shading, predicting the occluded surface necessitates a strong 3D object prior. Our field has seen a divergence in two different directions: feedforward regression~\cite{choy20163d,wang2018pixel2mesh,mescheder2019occupancy,xu2019disn,groueix2018,wu2017marrnet,huang2023shapeclipper,wu2023multiview,huang2024zeroshape,hong2023lrm,TripoSR2024,tang2024lgm,zhang2024gs,xu2024instantmesh,wang2024crm,sf3d2024} and diffusion-based generation~\cite{nichol2022point,jun2023shap,liu2023one2345++,shue20233d,chou2023diffusion,shim2023diffusion,cheng2023sdfusion,li2023diffusion,liu2023zero,shi2023mvdream,liu2023syncdreamer,huang2024pointinfinity,yariv2024mosaic,chen2023single,zhao2023michelangelo,lan2024ln3diff}. Despite the significant progress made in both directions, each has fundamental limitations.

Regression-based models are highly effective in adhering to the visible surface in the image, and the inference speed is typically fast. However, they make the oversimplified assumption of bijective mapping between images and 3D. This assumption introduces ambiguity in the learning objective, leading to poorly estimated surfaces and textures in occluded regions. On the other hand, diffusion-based approaches are generative and do not predict the statistical mean. However, their iterative sampling at inference time is computationally inefficient when modeling high-resolution 3D. Additionally, previous studies such as~\cite{huang2024zeroshape} indicate that diffusion-generated 3D models exhibit worse alignment to the surface visible in the input image. How can we take the best of both worlds while avoiding their limitations?

In light of this, we propose \method, which breaks the 3D reconstruction process down into two stages: the point sampling stage and the meshing stage. The point sampling stage uses diffusion models to generate sparse point clouds, followed by the meshing stage, transforming point clouds into highly detailed meshes. Our main idea is to offload the uncertainty modeling to the point sampling stage, where the low resolution of the point clouds allows rapid iterative sampling. The subsequent meshing stage leverages the local image features to transform the point cloud into a detailed mesh of high output fidelity. Reducing the meshing uncertainty with point clouds further facilitates unsupervised learning of inverse rendering, which reduces the baked-in lighting in the textures. Our two-stage design enables \method to significantly outperform previous regressive methods, while preserving high computational efficiency and fidelity to input observation.

A key design choice of our method is the usage of point clouds to connect the two stages. To ensure fast reconstruction, our intermediate representation needs to be lightweight so it can be efficiently generated. However, it should provide enough guidance to the meshing stage. This inspires us to use point clouds, which are perhaps the most computationally efficient 3D representation because all information bits are used to represent the surface. Moreover, the lack of connectivity, typically considered as the drawback of point clouds, now turns into an advantage with our two-stage approach for editing purposes. When the back surface does not align with user expectations, local edits can be easily made on the low-resolution point clouds without worrying about topologies (see~\cref{fig:teaser} bottom). Feeding edited point clouds into the meshing stage produces better meshes tailored towards user requirements.

Our experiments demonstrate the superiority of \method over previous state-of-the-art methods, with solid quantitative and qualitative results on various data sources.
\method also exhibits a strong generalization ability to in-the-wild images and AI-generated images. With a total inference time below \gentime seconds, \method is not only efficient but also allows for easy user-driven edits, offering a practical solution to the task of monocular 3D reconstruction. We hope that this is a meaningful step towards scalable generation of high-quality 3D assets.

\section{Related Work}
\label{sec:related}
\paragraph{Feedforward 3D reconstruction}
methods address the problem of 3D object reconstruction by learning a feedforward model in a regression-based manner.
Earlier works~\cite{choy20163d,wang2018pixel2mesh,mescheder2019occupancy,xu2019disn,groueix2018,wu2017marrnet,huang2023shapeclipper} in this field typically predict only the geometry and train on small datasets~\cite{chang2015shapenet,sun2018pix3d}, which limits their generalization ability. Recently, larger 3D datasets~\cite{reizenstein2021co3d,deitke2022objaverse} have been collected, unlocking the potential to train feedforward 3D models at scale~\cite{wu2023multiview,huang2024zeroshape,hong2023lrm}. These models exhibit great generalization ability to unseen images, and excel at producing reconstructions that tightly align with the observed cues in the input image. In particular, LRM~\cite{hong2023lrm} and follow-up works~\cite{TripoSR2024,tang2024lgm,zhang2024gs,xu2024instantmesh,wang2024crm,sf3d2024} show that properly designed large transformer models can be trained using only rendering losses to capture object geometry and texture in great detail. Despite the high fidelity and computational efficiency of these models, the oversimplified bijective assumption in these regressive approaches results in oversmoothed unseen surfaces. Multi-view diffusion models~\cite{liu2023zero,shi2023mvdream,liu2023syncdreamer,shi2023zero123plus} have been considered as a remedy for this, where additional viewpoints are synthesized as input to the feedforward model~\cite{tang2024lgm,wang2024crm,xu2024instantmesh}. However, the inconsistency across viewpoints often leads to significant artifacts on the reconstructed surfaces, and the computational efficiency of these approaches is severely affected by the slow multi-view generation process. Our model also aims to overcome the learning ambiguity in regressive approaches, but our point sampling approach is inherently 3D-consistent and computationally efficient, and further allows easy user edits.

\begin{figure*}[t]
\centering
\includegraphics[width=1.0\linewidth]{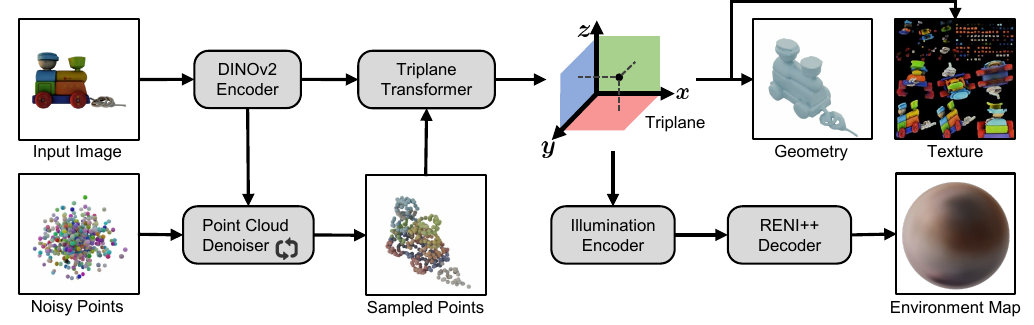}
\caption{\textbf{\method Overview.} Conditioned on the input image, \method first leverages a point diffusion model to generate a sparse point cloud. The triplane transformer then uses the sampled point cloud and image features to produce high-resolution triplane features. The triplane features are then queried to reconstruct the geometry, texture, and illumination of the object in the image.}
\vspace{-5mm}
\label{fig:overview}
\end{figure*}

\paragraph{Generative 3D modeling}
learns the image-conditioned distribution of 3D assets instead of a deterministic mapping. Early 3D generative works use GAN~\cite{valsesia2018learning,hui2020progressive,cai2020learning,gao2022get3d}, normalizing flow~\cite{yang2019pointflow,klokov2020discrete} or VAE~\cite{wu2019sagnet,mittal2022autosdf,gao2021tm} as the generative framework. Inspired by the success of 2D diffusion models~\cite{dhariwal2021diffusion,rombach2022high}, 3D diffusion models~\cite{nichol2022point,jun2023shap,liu2023one2345++,shue20233d,chou2023diffusion,shim2023diffusion,cheng2023sdfusion,li2023diffusion,liu2023zero,shi2023mvdream,liu2023syncdreamer,huang2024pointinfinity,yariv2024mosaic,chen2023single,zhao2023michelangelo,lan2024ln3diff} have also been extensively explored in recent works. Despite the advantage of probabilistic modeling that avoids over-smoothed results, diffusion-based 3D generation has two drawbacks: 1) not aligning well with input observations, and 2) having low inference speed at high resolution. Our work inherits the advantage of probabilistic modeling, while avoiding the drawbacks by using diffusion to generate only sparse point clouds.

\paragraph{Optimization-based single-view 3D} 
leverages 2D generative priors to recover 3D from single-view images. These works~\cite{melas2023realfusion,deng2023nerdi,tang2023make,gu2023nerfdiff} rely on SDS-type loss~\cite{pooledreamfusion, wang2023score} and generate 3D assets by optimizing for each object image separately. These methods achieve promising results without large-scale annotation. However, the lack of a strong explicit 3D prior makes the optimization process inefficient and prone to local minima.
\section{Method}
\label{sec:method}

\paragraph{\method Overview.}
Given the input image $I \in \mathbb{R}^{3 \times h \times w}$, our method produces a 3D mesh with PBR materials, including albedo, metallic, roughness and surface normals. The main goal of our work is to develop a model that enjoys the benefits of distribution learning through diffusion models, while not suffering from the low output fidelity and computational inefficiency. To this end, we design a two-stage model that consists of the point sampling stage and the meshing stage (see~\cref{fig:overview}). At the point sampling stage, a point diffusion model learns the conditional distribution of point clouds given the input image. This stage is computationally efficient given the low resolution of the point clouds. The regression-based meshing stage transforms the sampled point cloud into a highly detailed mesh that aligns with the visible surface. The reduced uncertainty with point sampling further facilitates the learning of materials and illumination in an unsupervised manner during the meshing stage. This reduces baked-in lighting artifacts and results in better modeling of specular surfaces. Finally, by using sparse point clouds as the intermediate representation, \method enables human editing in the loop.

\subsection{Point Sampling Stage}
\paragraph{Overview.}
The point sampling stage produces a sparse point cloud as the input to the meshing stage. The core of the point sampling stage is a point diffusion model, which generates point clouds $\point_0 \in \mathbb{R}^{n \times 6}$ conditioned on the input image $I$. The six channels include three XYZ channels and three RGB channels. In our work, the resolution of the point cloud $n$ is set to 512.

\paragraph{Point Diffusion Framework.}
Our diffusion framework is based on DDPM~\cite{ho2020denoising}, which consists of two processes: 1) the forward process which adds noise to the original point cloud, and 2) the backward process where the denoiser learns to remove the noise. At timestep $t \in [0, T]$, the diffusion process combines Gaussian noise $\boldsymbol{\epsilon} \sim \mathcal{N}(\mathbf{0},\mathbf{I})$ with a point cloud $\point_0$ as
\begin{equation}
    \point_t = \sqrt{\bar{\alpha_t}} \point_0 + \sqrt{1 - \bar{\alpha_t}} \boldsymbol{\epsilon},
\end{equation}
where $\bar{\alpha_t}$ denotes the noise schedule. We use the sigmoid noise schedule proposed in~\cite{chen2023importance}, combined with input scaling and the renormalization trick. The denoiser $\boldsymbol{\epsilon}_\theta (\point_t, t; \condition)$ then learns to recover the noise from $\point_t$ and is supervised by
\begin{equation}
    L_{simple}(\theta) = \mathbb{E}_{t,\point_0,\boldsymbol{\epsilon}} \lVert \boldsymbol{\epsilon} - \boldsymbol{\epsilon}_\theta (\point_t, t; \condition) \rVert_2^2.
\end{equation}
Here $\condition$ denotes the image condition tokens. During inference, we use the DDIM sampler~\cite{songdenoising} to generate point cloud samples. Samples generated directly often align poorly with the condition, hence we use the classifier-free guidance (CFG)~\cite{ho2022classifier} to improve sampling fidelity. 

\paragraph{Denoiser Design.}
We use a transformer denoiser similar to Point-E~\cite{nichol2022point}, where the noisy point cloud $\point_t \in \mathbb{R}^{n \times 6}$ is linearly mapped to a set of point tokens $\boldsymbol{x} \in \mathbb{R}^{n \times d}$. We use DINOv2~\cite{oquab2023dinov2} to encode the input image $I$ as conditioning tokens $\condition \in \mathbb{R}^{c \times d}$. The conditions and the point tokens are then concatenated together as input to the transformer, which predicts the added noise on each point. 

\paragraph{Albedo Point clouds.}
In the meshing stage, we estimate the materials and lighting alongside the geometry. However, this decomposition is inherently ambiguous because there are countless combinations of lighting and albedo that can explain the same input image. It is challenging to learn this highly uncertain decomposition during the regressive meshing stage alone. We therefore reduce the uncertainty at the point sampling stage, by directly generating albedo point clouds with diffusion models. Sampling albedo point clouds as input to the meshing stage drastically reduces the ambiguity of inverse rendering and stabilizes the decomposition learning.

\begin{figure*}[htb]
\begin{minipage}[t]{0.55\linewidth}
\centering
\includegraphics[width=1.0\linewidth]{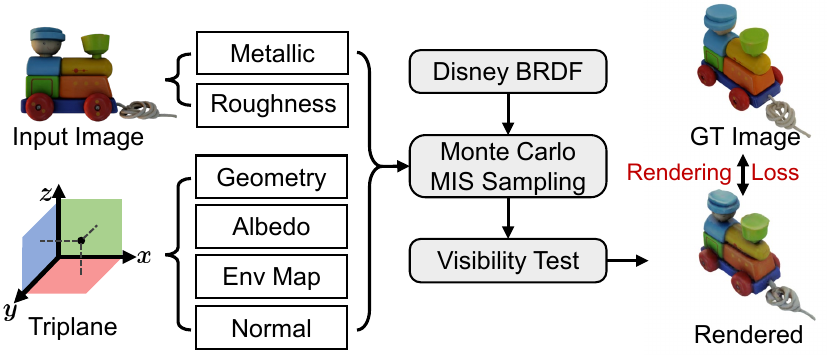}
\caption{\textbf{Our Differentiable Renderer.} We estimate geometry, albedo, lighting, and normal maps from the triplane and metallic/roughness values from the image. We rasterize and interpolate these values as input to our shader (omitted here for simplicity). Our shader uses the Disney BRDF~\cite{Burley2012} and performs Monte Carlo integration. We further perform visibility testing to improve shadow modeling. Finally, we compare the rendered image with the GT image and minimize the rendering loss.}
\label{fig:rendering}
\end{minipage}
\hfill
\begin{minipage}[t]{0.42\linewidth}
\centering
\includegraphics[width=\linewidth]{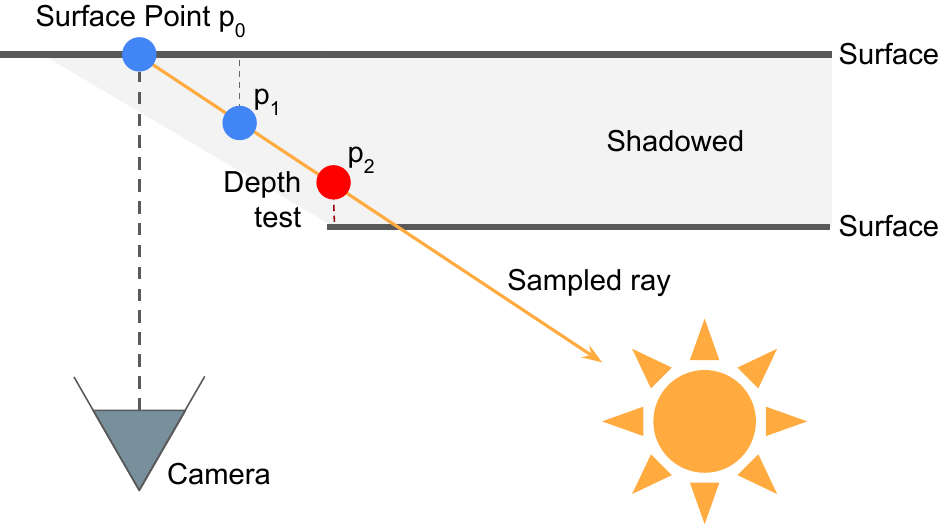}
\caption{\textbf{Shadow Modeling.} We perform visibility testing in screen-space by marching along sampled rays. If any point along the ray has a ray depth which is farther away than the depth map, we consider the entire ray as shadowed.}
\label{fig:shadow_test}
\end{minipage}
\vspace{-5mm}
\end{figure*}

\subsection{Meshing Stage}

\paragraph{Overview.}
The meshing stage produces a textured mesh from the input image and the point cloud. The backbone of our meshing model is a large triplane transformer, which predicts triplane features from the image and point cloud conditions. We estimate the geometry, texture and lighting of the current object from the triplane, and metallic/roughness from the image features. The geometry and materials are fed into our differentiable renderer during training, so that we can apply rendering loss to supervise our model.

\paragraph{Triplane Transformer.}
Our triplane transformer consists of three submodules: the point cloud encoder, the image encoder, and the transformer backbone. We use a simple transformer encoder to encode the point cloud as a set of point tokens. Given the low resolution of the point clouds, each point can be directly mapped to a single token. Our image encoder is DINOv2~\cite{oquab2023dinov2}, which produces local image embeddings. Our triplane transformer follows a similar design to PointInfinity~\cite{huang2024pointinfinity} and SF3D~\cite{sf3d2024}, which produces triplanes at high resolution of $384 \times 384$ by using a computationally-detached two-stream design.

\paragraph{Surface Estimation.}
To estimate the geometry, the triplanes are queried with a shallow MLP to produce density values. Similar to~\cite{sf3d2024,wang2024crm,xu2024instantmesh}, we convert the implicit density field to explicit surface using differentiable Marching Tetrahedron (DMTet)~\cite{shen2021deep}. We additionally use two MLP heads to predict vertex offsets and surface normals together with density. These two attributes reduce the artifacts introduced by the Marching Tetrahedron and lead to locally smoother surfaces.

\paragraph{Material and Illumination Estimation.}
We perform inverse rendering and jointly estimate materials (albedo, metallic and roughness) and illumination alongside the geometry. The task is highly ill-posed and Neural-PIL~\cite{Boss2021neuralPIL} showed that an illumination prior can reduce the ambiguity. We build our illumination estimator upon the learning-based illumination prior from RENI++~\cite{gardner2023reni++}. RENI++ is originally an unconditional generative model for HDR illumination generation. We learn an encoder to map triplane features into the latent space of RENI++. This allows us to estimate the environment illumination in the input image. The albedo is estimated from triplane similar to geometry, where a shallow MLP predicts the albedo value for each 3D location. For metallic and roughness, we follow SF3D~\cite{sf3d2024} and learn to estimate them with a probabilistic approach via a Beta prior. We find that the CLIP encoder used in SF3D is unstable when the object size changes. We therefore replace their CLIP encoder with AlphaCLIP~\cite{sun2023alphaclip} to alleviate this issue using foreground object masks.

\paragraph{Differentiable Rendering.}
We implement a differentiable renderer that renders images based on the predicted environment map, PBR materials and geometry surface (see~\cref{fig:rendering}). We use a differentiable mesh rasterizer and add a differentiable shader. Specifically, we leverage the standard simplified Disney PBR model~\cite{Burley2012} in our shader. As we use RENI++ to reconstruct environment maps, we need to explicitly integrate the incoming radiance. Here, we opt to use the Monte Carlo Integration. Given the low sample counts we can computationally afford during training, we rely on Multiple Importance Sampling (MIS) with the balanced heuristic~\cite{VeachPhD} to reduce integration variance. Additionally, to better model the self-occlusion which has been typically ignored in prior works, we implement a visibility test for better shadow modeling. We take inspiration from real-time graphics and model the visibility test as a screen-space method using the depth map from our rasterizer. An overview of this test is shown in Fig.~\ref{fig:shadow_test}. Specifically, we ray-march a short distance (0.25) in 6 steps for all proposed sample directions from MIS, and project the position back to image space. If the current ray depth is farther away than the sampled value from the depth map, then the ray is marked as shadowed. 

\paragraph{Loss Function.}
Our main loss function is the rendering loss that compares renderings from novel views to the grountruth (GT) images. Specifically, our rendering loss is a linear combination of 1) the L2 distance between the rendered and GT images, 2) the perceptual distance between the rendered and GT images measured by LPIPS~\cite{zhang2018perceptual}, and 3) the L2 distance between the rendered opacity and the GT foreground mask. Apart from the rendering loss, we also follow SF3D and apply the mesh and shading regularization that regularizes the surface smoothness and the inverse rendering respectively. 

\subsection{Interactive Editing}
A unique advantage of our two-stage design is that it naturally supports interactive editing of unseen regions in our produced mesh. In most circumstances, the visible surface is determined by the input image and remains highly accurate, while the unseen surface is mainly based on the sampled point cloud, which might not align with user intention. In this case, editing the unseen surface of the mesh is feasible by altering the point cloud. Point clouds are perhaps one of the most flexible 3D representation for editing purposes because there are no topology constraints. Given the low resolution of our point clouds, editing the point cloud is fairly efficient and intuitive. Users can easily delete, duplicate, stretch or recolor points in the point cloud. Our efficient meshing model is able to produce the adjusted mesh in 0.3 seconds, which makes this process fairly interactive. 

\subsection{Implementation Details}
\paragraph{Point Sampling Stage.}
Our point diffusion model has 16 transformer blocks in total. Each transformer block consists of two Layer Normalization layer, one Multi-Head Attention (MHA) layer and one MLP. We use a feature dimension of 1024 and 16 attention heads in each MHA layer. With many emissive objects in our dataset, albedo can be visually distinct from the input image and hard to learn. Therefore, instead of directly generating albedo point clouds in the point sampling stage, we learn to generate white-lit point clouds as a proxy target. 

\paragraph{Meshing Stage.}
Our triplane transformer consists of 4 two-stream blocks~\cite{huang2024pointinfinity}. Each two-stream block consists of three self-attentions and two cross-attentions. The main computation is carried out using 3,072 latent tokens, each with a feature dimension of 1024. The MHA includes 16 attention heads. The point cloud encoder is a vanilla transformer with 12 layers and 512 feature dimension, and the image encoder is DINOv2-large. We use a tetrahedra resolution of 160 for DMTet. In our differentiable shader, we follow Hasselgren \etal~\cite{hasselgren2022nvdiffrecmc} and use a detached biased sampling scheme. We sample based on the specular lobe (GGX~\cite{Walter2007}), the 2D piecewise-linear distribution of the environment map luminance and the hemispherical distribution.  Specifically, we include 6 samples from the GGX lobe, 6 samples from the 2D piecewise-linear distribution of the luminance, and 4 samples from the hemispherical distribution. The main body of the shader is implemented in PyTorch, while the screen-space shadowing and the 2D piecewise-linear distribution computation for the environment map are implemented as custom CUDA kernels for efficiency. The training of our meshing stage includes multiple phases, where we increase the rendering resolution and decrease the batch size at later training phases. We use GT point clouds as input when training the meshing model. The curation of our training data follows TripoSR~\cite{TripoSR2024}. 
\section{Experiments}
\label{sec:experiments}

\begin{table*}[h]
\begin{center}
\scriptsize
\resizebox{1.0\linewidth}{!}{\begin{tabular}{ l c c c c c c c c}
\hline
Method & CD$\downarrow$ & FS@0.1$\uparrow$ & FS@0.2$\uparrow$ & FS@0.5$\uparrow$ & PSNR$\uparrow$ & SSIM$\uparrow$ & LPIPS$\downarrow$ & Time (s)$\downarrow$ \\
\hline
Shap-E~\cite{jun2023shap}               & 0.204 & 0.359 & 0.638 & 0.922 & 15.3 & 0.802 & 0.205 & 3.1    \\
LN3Diff~\cite{lan2024ln3diff}              & 0.174 & 0.422 & 0.703 & 0.949 & 17.1 & 0.819 & 0.169 & 5.1    \\
LGM~\cite{tang2024lgm}                  & 0.196 & 0.356 & 0.635 & 0.936 & 17.0 & 0.818 & 0.184 & 41.0    \\
CRM~\cite{wang2024crm}                  & 0.161 & 0.437 & 0.735 & 0.961 & 17.5 & 0.830 & 0.169 & 7.4    \\
TripoSR~\cite{TripoSR2024}              & 0.145 & 0.501 & 0.784 & 0.968 & \underline{18.5} & \underline{0.837} & 0.151 & \textbf{0.2}    \\
InstantMesh~\cite{xu2024instantmesh}    & \underline{0.135} & \underline{0.545} & \underline{0.812} & \underline{0.971} & 18.1 & \underline{0.838} & \underline{0.146} & 36.1    \\
SF3D~\cite{sf3d2024}                    & \underline{0.137} & \underline{0.540} & \underline{0.806} & \underline{0.970} & 18.0 & \textbf{0.839} & \underline{0.145} & \underline{0.3}    \\
\method (ours)                          & \textbf{0.120} & \textbf{0.584} & \textbf{0.850} & \textbf{0.983} & \textbf{18.6} & \underline{0.836} & \textbf{0.139} & \underline{0.7}    \\
\hline
\end{tabular}}
\caption{\textbf{Quantitative Comparisons on GSO~\cite{downs2022google}.} \method performs favorably to other state-of-the-art methods.}
\label{tab:gso}
\vspace{-3mm}
\end{center}
\end{table*}
\begin{table*}[h]
\begin{center}
\scriptsize
\resizebox{1.0\linewidth}{!}{\begin{tabular}{l c c c c c c c c}
\hline
Method & CD$\downarrow$ & FS@0.1$\uparrow$ & FS@0.2$\uparrow$ & FS@0.5$\uparrow$ & PSNR$\uparrow$ & SSIM$\uparrow$ & LPIPS$\downarrow$ & Time (s)$\downarrow$ \\
\hline
Shap-E~\cite{jun2023shap}               & 0.212 & 0.349 & 0.624 & 0.909 & 14.8 & 0.8006 & 0.205 & 3.1    \\
LN3Diff~\cite{lan2024ln3diff}              & 0.160 & 0.480 & 0.744 & 0.957 & 16.7 & 0.819 & 0.161 & 5.0    \\
LGM~\cite{tang2024lgm}                  & 0.200 & 0.366 & 0.638 & 0.924 & 16.1 & 0.810 & 0.188 & 42.0    \\
CRM~\cite{wang2024crm}                  & 0.155 & 0.482 & 0.765 & 0.962 & 17.0 & 0.828 & 0.162 & 7.0    \\
TripoSR~\cite{TripoSR2024}              & 0.144 & 0.537 & 0.785 & 0.963 & \textbf{18.0} & \underline{0.835} & 0.147 & \textbf{0.2}    \\
InstantMesh~\cite{xu2024instantmesh}    & 0.145 & 0.546 & 0.790 & 0.962 & 17.2 & 0.832 & 0.150 & 34.7    \\
SF3D~\cite{sf3d2024}                    & \underline{0.138} & \underline{0.554} & \underline{0.800} & \underline{0.967} & 17.4 & \textbf{0.836} & \underline{0.145} & \underline{0.3}    \\
\method (ours)                          & \textbf{0.122} & \textbf{0.587} & \textbf{0.845} & \textbf{0.978} & \underline{17.9} & 0.832 & \textbf{0.140} & \underline{0.7}    \\
\hline
\end{tabular}}
\caption{\textbf{Quantitative Comparisons on OmniObject3D~\cite{wu2023omniobject3d}.} \method performs favorably to other state-of-the-art methods.}
\label{tab:omni}
\vspace{-8mm}
\end{center}
\end{table*}

\subsection{Evaluation}
\paragraph{Datasets.}
We used two datasets for evaluation, GSO~\cite{downs2022google} and OmniObject3D~\cite{wu2023omniobject3d}. We follow TripoSR~\cite{TripoSR2024} and remove simple box or cylindrical objects to avoid bias on simple geometries. Each of the evaluation sets consists of around 250 objects. We render the objects with diverse azimuth angles at different elevations, with randomly sampled HDRI environment maps. We also vary the focal length of the camera to create more diverse test cases.

\paragraph{Metrics.}
To evaluate the geometry quality of the reconstructed meshes, we use follow prior works~\cite{TripoSR2024,xu2024instantmesh} and use Chamfer Distance (CD) and F-score (FS) as our evaluation metrics. CD measures the alignment between two point clouds and is defined as the average of accuracy and completeness:
\begin{equation}
    \small
    d(S_1, S_2) = \frac{1}{2|S_1|}\sum_{x \in S_1} \min_{y \in S_2} \|x-y\|_2 + \frac{1}{2|S_2|}\sum_{y \in S_2} \min_{x \in S_1} \|x-y\|_2
    \label{eq:chamfer}
\end{equation}
FS evaluates point cloud alignment by calculating the F-score with a predefined threshold. Predicted points that lie within the distance threshold are considered as correct predictions. A higher FS means better alignment between the reconstructed shape and the groundtruth.
To evaluate the texture quality, we compute standard image metrics, including PSNR, SSIM and LPIPS, between images rendered from the predicted mesh and the groundtruth images.

\begin{figure*}[t]
\centering
\includegraphics[width=0.95\textwidth]{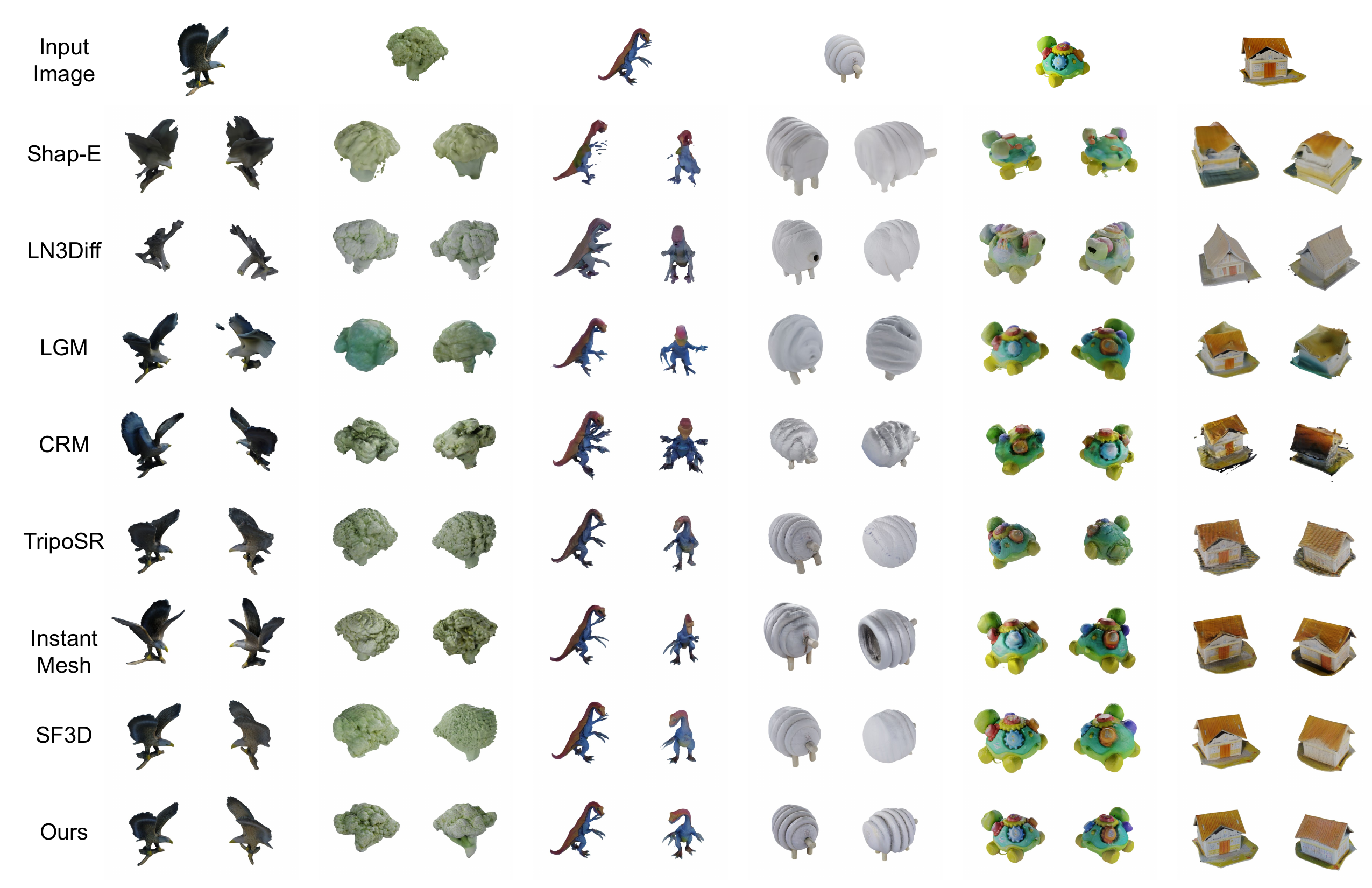}
\vspace{-4mm}
\caption{\textbf{Qualitative Comparison.} We compare \method to other state-of-the-art methods visually. \method not only aligns better with the visible surfaces from images, but also generates higher-quality geometries and textures for the occluded surfaces.}
\label{fig:qualitative}
\vspace{-4mm}
\end{figure*}

\paragraph{Protocol.}
To calculate the metrics that are comparable across methods, the meshes need to lie in the same coordinate system. To this end, we perform brute-force search in rotations to align each predicted mesh with the groundtruth mesh. Both the prediction and the groundtruth are normalized before the brute-force alignment, and the alignment is further refined with ICP.

\paragraph{Baselines.}
We compare \method with other efficient methods for single-view 3D generation or reconstruction~\cite{TripoSR2024,tang2024lgm,xu2024instantmesh,wang2024crm,sf3d2024}. We use the official implementation for all baselines, and we evaluate the produced meshes under the same protocol. Specifically, we compare against TripoSR~\cite{TripoSR2024}, LGM~\cite{tang2024lgm}, CRM~\cite{wang2024crm}, InstantMesh~\cite{xu2024instantmesh}, LN3Diff~\cite{lan2024ln3diff}, Shap-E~\cite{jun2023shap} and SF3D~\cite{sf3d2024}. Among these baselines, TripoSR and SF3D are pure regression-based approaches; LGM, CRM and InstantMesh use multiview diffusion to generate pseudo multi-view images; LN3Diff and Shap-E are purely diffusion-based 3D generative models.

\begin{figure*}[h]
\centering
\includegraphics[width=0.95\textwidth]{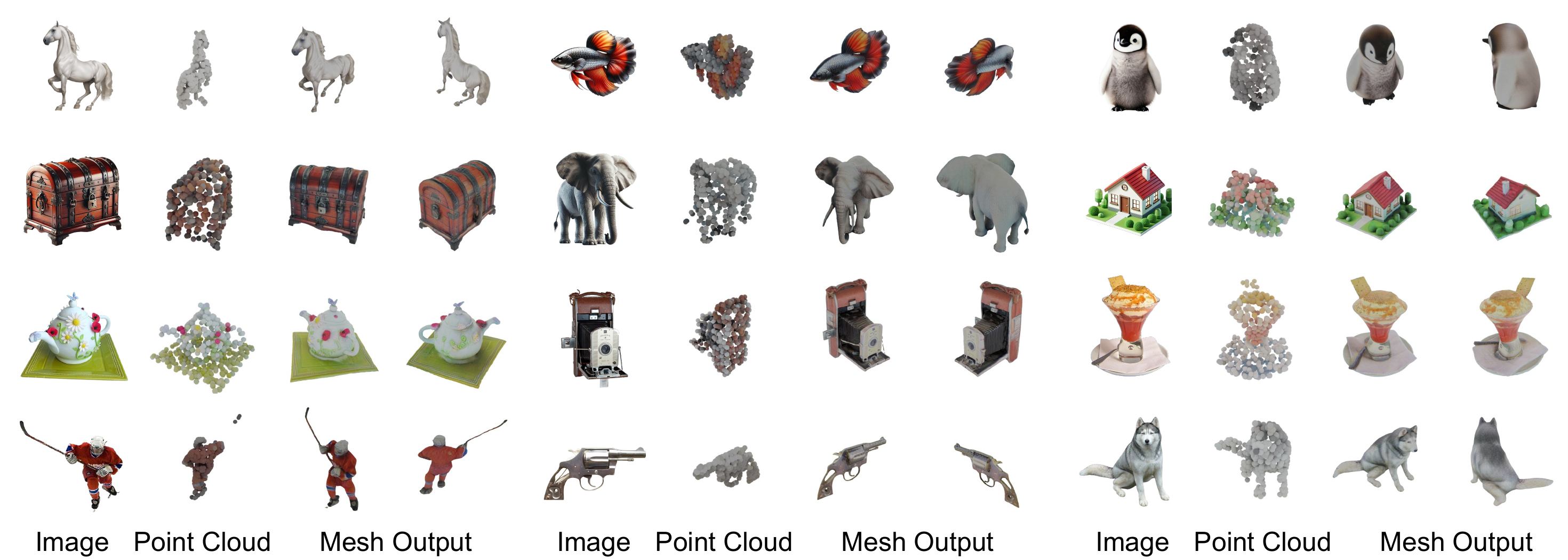}
\caption{\textbf{Generalization Results.} We show qualitative results of \method on in-the-wild images from 2D generative models (top 2 rows) and ImageNet (bottom 2 rows). The reconstructed meshes exhibit accurate geometric structures with great textures, demonstrating a strong generalization performance of \method.}
\vspace{-6mm}
\label{fig:wild}
\end{figure*}

\begin{figure*}[h]
\centering
\includegraphics[width=0.95\textwidth]{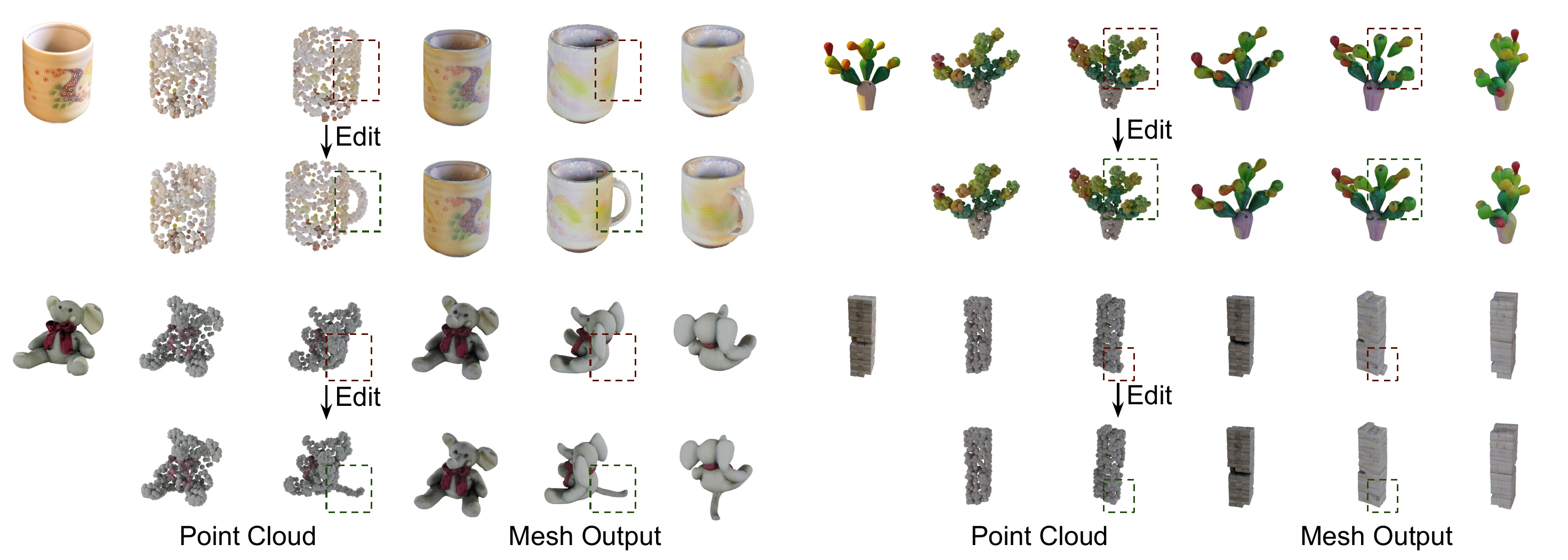}
\caption{\textbf{Editing Results.} We show qualitative examples of interactive editing with \method. On the left two examples, we add a handle to the mug and a tail to the elephant doll by duplicating existing points. On the right two examples, we move or delete points to fix imperfections and to improve local details on the mesh. All the edits are performed in Blender within a minute.}
\label{fig:edit}
\vspace{-6mm}
\end{figure*}

\subsection{Main Results}
\paragraph{Quantitative Comparison.}
We compare \method to other baselines on GSO and Omniobject3D quantitatively. As shown in~\cref{tab:gso} and~\cref{tab:omni}, \method outperforms all other regressive or generative baselines significantly across most metrics on both datasets. For SSIM, we observe that \method is slightly worse than the strongest baseline for this metric. We find that this relates to the Monte Carlo noise from our shader. \method is also among the fastest reconstruction models with an inference speed of \gentime seconds per object, which is significantly faster than 3D or multiview diffusion-based approaches. 

\paragraph{Qualitative Results.}
We show qualitative results of different methods in~\cref{fig:qualitative}. The reconstructed meshes from pure regression-based approaches such as SF3D or TripoSR align with the input image well, but the backside is often less accurate and over-smoothed. Multi-view diffusion-based methods such as LGM, CRM and InstantMesh show more details on the backside. However, the inconsistency in the synthesized views leads to clear artifacts and overall worse results. Pure generative approaches such as Shap-E and LN3Diff are able to produce sharp surfaces in their generation. However, many details are erroneous hallucinations that do not accurately follow the input images, and the visible surfaces are often reconstructed incorrectly. Compared to prior art, the meshes produced by \method not only faithfully resemble the input image, but also exhibit well-generated occluded parts with reasonable details. In~\cref{fig:wild}, we further show qualitative results of \method on in-the-wild images. The images are generated using text-to-image generative models or from the ImageNet validation set~\cite{deng2009imagenet}. The high quality of the reconstructed meshes demonstrates a strong generalization performance of \method.

\subsection{Editing Results}
The usage of explicit point clouds as an intermediate representation enables interactive editing of the generated meshes. Users can easily alter the unseen surface of the mesh by manipulating the point cloud. In~\cref{fig:edit}, we show a few editing examples with \method, either by adding major object parts to the reconstruction, or improving undesirable generated details.

\subsection{Ablation}
We ablate the key idea of \method, the point sampling stage, which can be seen as an addition to standard regression approaches. We consider a variant of our model (\method w/o Point), where we remove the point sampling stage and make \method a full regressive model.
We compare this variant with our full model on both GSO and Omniobject3D.
As shown in~\cref{tab:ablation}, our full \method significantly outperforms the regressive variant, which validates the effectiveness of our design.

\begin{table}[htb]
\begin{center}
\scriptsize
\resizebox{1.0\linewidth}{!}{\begin{tabular}{l c c c c}
\hline
Method & CD$\downarrow$ & FS@0.1$\uparrow$ & PSNR$\uparrow$ & LPIPS$\downarrow$ \\
\hline
\method w/o Point                   & 0.136 & 0.506 & 18.5 & 0.146    \\
\method                             & \textbf{0.120} & \textbf{0.584} & \textbf{18.6} & \textbf{0.139}    \\
\hline
\hline
\method w/o Point                   & 0.140 & 0.509 & 17.8 & 0.146    \\
\method                             & \textbf{0.122} & \textbf{0.587} & \textbf{17.9} & \textbf{0.140}    \\
\hline
\end{tabular}}
\caption{\textbf{Ablation Study on GSO (top 2 rows) and Omniobject3D (bottom 2 rows).} Removing the point sampling stage leads to significant performance drop.}
\label{tab:ablation}
\end{center}
\end{table}
\vspace{-6mm}

\begin{figure}[h]
\centering
\includegraphics[width=0.9\linewidth]{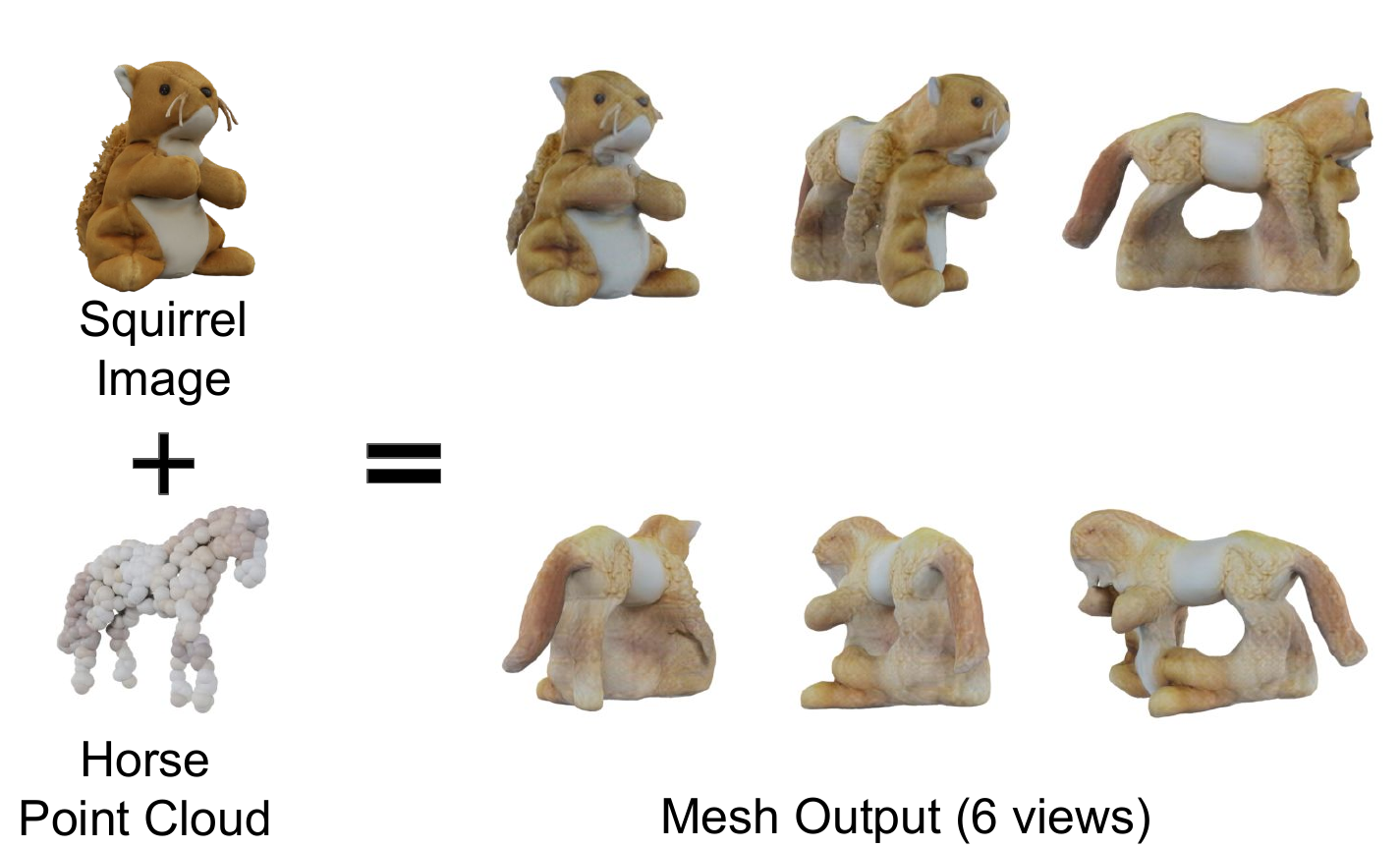}
\caption{\textbf{Generated Mesh with Conflicting Cues.} Under conflicting cues from images and point clouds, our model reconstructs the visible surface based on the image, while generating the backside surface based on the point cloud.}
\label{fig:analysis}
\vspace{-5mm}
\end{figure}

\subsection{Analysis}
We further design experiments to understand how \method works. 
Our key assumption when designing \method is that the two-stage design effectively separates the uncertain part (back-surface modeling) and the deterministic part (visible surface modeling) of the monocular 3D reconstruction problem. Ideally, the meshing stage should mainly rely on the input image for reconstructing the visible surface, while relying on the point cloud to generate the back surface. To see whether this is true, we design an experiment where we artificially use point clouds that conflict with the input image. In~\cref{fig:analysis}, we feed the input image of a squirrel and the point cloud of a horse to the meshing model. As shown in the figure, the reconstructed mesh indeed aligns with the squirrel image well on the visible surface, while the back surface mainly adheres to the point cloud. This result validates our assumption. 
\section{Conclusion}
\label{sec:conclusion}

We present \method, a simple yet effective approach for single-view 3D reconstruction. The core of our model is a two-stage design based on point sampling. We first generate a sparse point cloud via point diffusion, and then reconstruct a highly detailed mesh from both the point cloud and the image. This design enables us to take the best of regression-based and generative modeling. Evaluated on standard benchmarks and in-the-wild images, \method significantly outperforms previous state-of-the-art methods with a fast inference speed. We will release our model upon publication, and we hope our effort is useful for future research towards scalable generation of high-quality 3D content.
{
    \small
    \bibliographystyle{ieeenat_fullname}
    \bibliography{main}

\begin{thebibliography}{71}
\providecommand{\natexlab}[1]{#1}
\providecommand{\url}[1]{\texttt{#1}}
\expandafter\ifx\csname urlstyle\endcsname\relax
  \providecommand{\doi}[1]{doi: #1}\else
  \providecommand{\doi}{doi: \begingroup \urlstyle{rm}\Url}\fi

\bibitem[Boss et~al.(2021)Boss, Jampani, Braun, Liu, Barron, and Lensch]{Boss2021neuralPIL}
Mark Boss, Varun Jampani, Raphael Braun, Ce Liu, Jonathan~T. Barron, and Hendrik~P.A. Lensch.
\newblock Neural-pil: Neural pre-integrated lighting for reflectance decomposition.
\newblock \emph{NeurIPS}, 2021.

\bibitem[Boss et~al.(2024)Boss, Huang, Vasishta, and Jampani]{sf3d2024}
Mark Boss, Zixuan Huang, Aaryaman Vasishta, and Varun Jampani.
\newblock Sf3d: Stable fast 3d mesh reconstruction with uv-unwrapping and illumination disentanglement.
\newblock \emph{arXiv preprint}, 2024.

\bibitem[Burley(2012)]{Burley2012}
Brent Burley.
\newblock Physically-based shading at disney.
\newblock \emph{ACM Transactions on Graphics (SIGGRAPH)}, 2012.

\bibitem[Cai et~al.(2020)Cai, Yang, Averbuch-Elor, Hao, Belongie, Snavely, and Hariharan]{cai2020learning}
Ruojin Cai, Guandao Yang, Hadar Averbuch-Elor, Zekun Hao, Serge Belongie, Noah Snavely, and Bharath Hariharan.
\newblock Learning gradient fields for shape generation.
\newblock In \emph{Computer Vision--ECCV 2020: 16th European Conference, Glasgow, UK, August 23--28, 2020, Proceedings, Part III 16}, pages 364--381. Springer, 2020.

\bibitem[Chang et~al.(2015)Chang, Funkhouser, Guibas, Hanrahan, Huang, Li, Savarese, Savva, Song, Su, et~al.]{chang2015shapenet}
Angel~X Chang, Thomas Funkhouser, Leonidas Guibas, Pat Hanrahan, Qixing Huang, Zimo Li, Silvio Savarese, Manolis Savva, Shuran Song, Hao Su, et~al.
\newblock Shapenet: An information-rich 3d model repository.
\newblock \emph{arXiv preprint arXiv:1512.03012}, 2015.

\bibitem[Chen et~al.(2023)Chen, Gu, Chen, Tian, Tu, Liu, and Su]{chen2023single}
Hansheng Chen, Jiatao Gu, Anpei Chen, Wei Tian, Zhuowen Tu, Lingjie Liu, and Hao Su.
\newblock Single-stage diffusion nerf: A unified approach to 3d generation and reconstruction.
\newblock In \emph{Proceedings of the IEEE/CVF international conference on computer vision}, pages 2416--2425, 2023.

\bibitem[Chen(2023)]{chen2023importance}
Ting Chen.
\newblock On the importance of noise scheduling for diffusion models.
\newblock \emph{arXiv preprint arXiv:2301.10972}, 2023.

\bibitem[Cheng et~al.(2023)Cheng, Lee, Tulyakov, Schwing, and Gui]{cheng2023sdfusion}
Yen-Chi Cheng, Hsin-Ying Lee, Sergey Tulyakov, Alexander~G Schwing, and Liang-Yan Gui.
\newblock Sdfusion: Multimodal 3d shape completion, reconstruction, and generation.
\newblock In \emph{Proceedings of the IEEE/CVF Conference on Computer Vision and Pattern Recognition}, pages 4456--4465, 2023.

\bibitem[Chou et~al.(2023)Chou, Bahat, and Heide]{chou2023diffusion}
Gene Chou, Yuval Bahat, and Felix Heide.
\newblock Diffusion-sdf: Conditional generative modeling of signed distance functions.
\newblock In \emph{Proceedings of the IEEE/CVF International Conference on Computer Vision}, pages 2262--2272, 2023.

\bibitem[Choy et~al.(2016)Choy, Xu, Gwak, Chen, and Savarese]{choy20163d}
Christopher~B Choy, Danfei Xu, JunYoung Gwak, Kevin Chen, and Silvio Savarese.
\newblock 3d-r2n2: A unified approach for single and multi-view 3d object reconstruction.
\newblock In \emph{European conference on computer vision}, pages 628--644. Springer, 2016.

\bibitem[Deitke et~al.(2022)Deitke, Schwenk, Salvador, Weihs, Michel, VanderBilt, Schmidt, Ehsani, Kembhavi, and Farhadi]{deitke2022objaverse}
Matt Deitke, Dustin Schwenk, Jordi Salvador, Luca Weihs, Oscar Michel, Eli VanderBilt, Ludwig Schmidt, Kiana Ehsani, Aniruddha Kembhavi, and Ali Farhadi.
\newblock Objaverse: A universe of annotated 3d objects.
\newblock \emph{arXiv preprint arXiv:2212.08051}, 2022.

\bibitem[Deng et~al.(2023)Deng, Jiang, Qi, Yan, Zhou, Guibas, Anguelov, et~al.]{deng2023nerdi}
Congyue Deng, Chiyu Jiang, Charles~R Qi, Xinchen Yan, Yin Zhou, Leonidas Guibas, Dragomir Anguelov, et~al.
\newblock Nerdi: Single-view nerf synthesis with language-guided diffusion as general image priors.
\newblock In \emph{Proceedings of the IEEE/CVF conference on computer vision and pattern recognition}, pages 20637--20647, 2023.

\bibitem[Deng et~al.(2009)Deng, Dong, Socher, Li, Li, and Fei-Fei]{deng2009imagenet}
Jia Deng, Wei Dong, Richard Socher, Li-Jia Li, Kai Li, and Li Fei-Fei.
\newblock Imagenet: A large-scale hierarchical image database.
\newblock In \emph{2009 IEEE conference on computer vision and pattern recognition}, pages 248--255. Ieee, 2009.

\bibitem[Dhariwal and Nichol(2021)]{dhariwal2021diffusion}
Prafulla Dhariwal and Alexander Nichol.
\newblock Diffusion models beat gans on image synthesis.
\newblock \emph{Advances in neural information processing systems}, 34:\penalty0 8780--8794, 2021.

\bibitem[Downs et~al.(2022)Downs, Francis, Koenig, Kinman, Hickman, Reymann, McHugh, and Vanhoucke]{downs2022google}
Laura Downs, Anthony Francis, Nate Koenig, Brandon Kinman, Ryan Hickman, Krista Reymann, Thomas~B McHugh, and Vincent Vanhoucke.
\newblock Google scanned objects: A high-quality dataset of 3d scanned household items.
\newblock In \emph{2022 International Conference on Robotics and Automation (ICRA)}, pages 2553--2560. IEEE, 2022.

\bibitem[Gao et~al.(2022)Gao, Shen, Wang, Chen, Yin, Li, Litany, Gojcic, and Fidler]{gao2022get3d}
Jun Gao, Tianchang Shen, Zian Wang, Wenzheng Chen, Kangxue Yin, Daiqing Li, Or Litany, Zan Gojcic, and Sanja Fidler.
\newblock Get3d: A generative model of high quality 3d textured shapes learned from images.
\newblock \emph{Advances In Neural Information Processing Systems}, 35:\penalty0 31841--31854, 2022.

\bibitem[Gao et~al.(2021)Gao, Wu, Yuan, Lin, Lai, and Zhang]{gao2021tm}
Lin Gao, Tong Wu, Yu-Jie Yuan, Ming-Xian Lin, Yu-Kun Lai, and Hao Zhang.
\newblock Tm-net: Deep generative networks for textured meshes.
\newblock \emph{ACM Transactions on Graphics (TOG)}, 40\penalty0 (6):\penalty0 1--15, 2021.

\bibitem[Gardner et~al.(2023)Gardner, Egger, and Smith]{gardner2023reni++}
James~AD Gardner, Bernhard Egger, and William~AP Smith.
\newblock Reni++ a rotation-equivariant, scale-invariant, natural illumination prior.
\newblock \emph{arXiv preprint arXiv:2311.09361}, 2023.

\bibitem[Groueix et~al.(2018)Groueix, Fisher, Kim, Russell, and Aubry]{groueix2018}
Thibault Groueix, Matthew Fisher, Vladimir~G. Kim, Bryan Russell, and Mathieu Aubry.
\newblock {AtlasNet: A Papier-M\^ach\'e Approach to Learning 3D Surface Generation}.
\newblock In \emph{Proceedings IEEE Conf. on Computer Vision and Pattern Recognition (CVPR)}, 2018.

\bibitem[Gu et~al.(2023)Gu, Trevithick, Lin, Susskind, Theobalt, Liu, and Ramamoorthi]{gu2023nerfdiff}
Jiatao Gu, Alex Trevithick, Kai-En Lin, Joshua~M Susskind, Christian Theobalt, Lingjie Liu, and Ravi Ramamoorthi.
\newblock Nerfdiff: Single-image view synthesis with nerf-guided distillation from 3d-aware diffusion.
\newblock In \emph{International Conference on Machine Learning}, pages 11808--11826. PMLR, 2023.

\bibitem[Hasselgren et~al.(2022)Hasselgren, Hofmann, and Munkberg]{hasselgren2022nvdiffrecmc}
Jon Hasselgren, Nikolai Hofmann, and Jacob Munkberg.
\newblock {Shape, Light, and Material Decomposition from Images using Monte Carlo Rendering and Denoising}.
\newblock \emph{Adv. Neural Inform. Process. Syst. (NeurIPS)}, 2022.

\bibitem[Ho and Salimans(2022)]{ho2022classifier}
Jonathan Ho and Tim Salimans.
\newblock Classifier-free diffusion guidance.
\newblock \emph{arXiv preprint arXiv:2207.12598}, 2022.

\bibitem[Ho et~al.(2020)Ho, Jain, and Abbeel]{ho2020denoising}
Jonathan Ho, Ajay Jain, and Pieter Abbeel.
\newblock Denoising diffusion probabilistic models.
\newblock \emph{Advances in neural information processing systems}, 33:\penalty0 6840--6851, 2020.

\bibitem[Hong et~al.(2023)Hong, Zhang, Gu, Bi, Zhou, Liu, Liu, Sunkavalli, Bui, and Tan]{hong2023lrm}
Yicong Hong, Kai Zhang, Jiuxiang Gu, Sai Bi, Yang Zhou, Difan Liu, Feng Liu, Kalyan Sunkavalli, Trung Bui, and Hao Tan.
\newblock Lrm: Large reconstruction model for single image to 3d.
\newblock \emph{arXiv preprint arXiv:2311.04400}, 2023.

\bibitem[Huang et~al.(2023)Huang, Jampani, Thai, Li, Stojanov, and Rehg]{huang2023shapeclipper}
Zixuan Huang, Varun Jampani, Anh Thai, Yuanzhen Li, Stefan Stojanov, and James~M Rehg.
\newblock Shapeclipper: Scalable 3d shape learning from single-view images via geometric and clip-based consistency.
\newblock In \emph{Proceedings of the IEEE/CVF Conference on Computer Vision and Pattern Recognition}, 2023.

\bibitem[Huang et~al.(2024{\natexlab{a}})Huang, Johnson, Debnath, Rehg, and Wu]{huang2024pointinfinity}
Zixuan Huang, Justin Johnson, Shoubhik Debnath, James~M Rehg, and Chao-Yuan Wu.
\newblock Pointinfinity: Resolution-invariant point diffusion models.
\newblock In \emph{Proceedings of the IEEE/CVF Conference on Computer Vision and Pattern Recognition}, pages 10050--10060, 2024{\natexlab{a}}.

\bibitem[Huang et~al.(2024{\natexlab{b}})Huang, Stojanov, Thai, Jampani, and Rehg]{huang2024zeroshape}
Zixuan Huang, Stefan Stojanov, Anh Thai, Varun Jampani, and James~M Rehg.
\newblock Zeroshape: Regression-based zero-shot shape reconstruction.
\newblock In \emph{Proceedings of the IEEE/CVF Conference on Computer Vision and Pattern Recognition}, pages 10061--10071, 2024{\natexlab{b}}.

\bibitem[Hui et~al.(2020)Hui, Xu, Xie, Qian, and Yang]{hui2020progressive}
Le Hui, Rui Xu, Jin Xie, Jianjun Qian, and Jian Yang.
\newblock Progressive point cloud deconvolution generation network.
\newblock In \emph{Computer Vision--ECCV 2020: 16th European Conference, Glasgow, UK, August 23--28, 2020, Proceedings, Part XV 16}, pages 397--413. Springer, 2020.

\bibitem[Jun and Nichol(2023)]{jun2023shap}
Heewoo Jun and Alex Nichol.
\newblock Shap-e: Generating conditional 3d implicit functions.
\newblock \emph{arXiv preprint arXiv:2305.02463}, 2023.

\bibitem[Klokov et~al.(2020)Klokov, Boyer, and Verbeek]{klokov2020discrete}
Roman Klokov, Edmond Boyer, and Jakob Verbeek.
\newblock Discrete point flow networks for efficient point cloud generation.
\newblock In \emph{European Conference on Computer Vision}, pages 694--710. Springer, 2020.

\bibitem[Lan et~al.(2024)Lan, Hong, Yang, Zhou, Meng, Dai, Pan, and Loy]{lan2024ln3diff}
Yushi Lan, Fangzhou Hong, Shuai Yang, Shangchen Zhou, Xuyi Meng, Bo Dai, Xingang Pan, and Chen~Change Loy.
\newblock Ln3diff: Scalable latent neural fields diffusion for speedy 3d generation.
\newblock \emph{arXiv preprint arXiv:2403.12019}, 2024.

\bibitem[Li et~al.(2023)Li, Duan, Zhou, and Lu]{li2023diffusion}
Muheng Li, Yueqi Duan, Jie Zhou, and Jiwen Lu.
\newblock Diffusion-sdf: Text-to-shape via voxelized diffusion.
\newblock In \emph{Proceedings of the IEEE/CVF Conference on Computer Vision and Pattern Recognition}, pages 12642--12651, 2023.

\bibitem[Liu et~al.(2023{\natexlab{a}})Liu, Shi, Chen, Zhang, Xu, Wei, Chen, Zeng, Gu, and Su]{liu2023one2345++}
Minghua Liu, Ruoxi Shi, Linghao Chen, Zhuoyang Zhang, Chao Xu, Xinyue Wei, Hansheng Chen, Chong Zeng, Jiayuan Gu, and Hao Su.
\newblock One-2-3-45++: Fast single image to 3d objects with consistent multi-view generation and 3d diffusion.
\newblock \emph{arXiv preprint arXiv:2311.07885}, 2023{\natexlab{a}}.

\bibitem[Liu et~al.(2023{\natexlab{b}})Liu, Wu, Van~Hoorick, Tokmakov, Zakharov, and Vondrick]{liu2023zero}
Ruoshi Liu, Rundi Wu, Basile Van~Hoorick, Pavel Tokmakov, Sergey Zakharov, and Carl Vondrick.
\newblock Zero-1-to-3: Zero-shot one image to 3d object.
\newblock In \emph{Proceedings of the IEEE/CVF International Conference on Computer Vision}, pages 9298--9309, 2023{\natexlab{b}}.

\bibitem[Liu et~al.(2023{\natexlab{c}})Liu, Lin, Zeng, Long, Liu, Komura, and Wang]{liu2023syncdreamer}
Yuan Liu, Cheng Lin, Zijiao Zeng, Xiaoxiao Long, Lingjie Liu, Taku Komura, and Wenping Wang.
\newblock Syncdreamer: Generating multiview-consistent images from a single-view image.
\newblock \emph{arXiv preprint arXiv:2309.03453}, 2023{\natexlab{c}}.

\bibitem[Melas-Kyriazi et~al.(2023)Melas-Kyriazi, Laina, Rupprecht, and Vedaldi]{melas2023realfusion}
Luke Melas-Kyriazi, Iro Laina, Christian Rupprecht, and Andrea Vedaldi.
\newblock Realfusion: 360deg reconstruction of any object from a single image.
\newblock In \emph{Proceedings of the IEEE/CVF conference on computer vision and pattern recognition}, pages 8446--8455, 2023.

\bibitem[Mescheder et~al.(2019)Mescheder, Oechsle, Niemeyer, Nowozin, and Geiger]{mescheder2019occupancy}
Lars Mescheder, Michael Oechsle, Michael Niemeyer, Sebastian Nowozin, and Andreas Geiger.
\newblock Occupancy networks: Learning 3d reconstruction in function space.
\newblock In \emph{Proceedings of the IEEE/CVF Conference on Computer Vision and Pattern Recognition}, pages 4460--4470, 2019.

\bibitem[Mittal et~al.(2022)Mittal, Cheng, Singh, and Tulsiani]{mittal2022autosdf}
Paritosh Mittal, Yen-Chi Cheng, Maneesh Singh, and Shubham Tulsiani.
\newblock Autosdf: Shape priors for 3d completion, reconstruction and generation.
\newblock In \emph{Proceedings of the IEEE/CVF Conference on Computer Vision and Pattern Recognition}, pages 306--315, 2022.

\bibitem[Nichol et~al.(2022)Nichol, Jun, Dhariwal, Mishkin, and Chen]{nichol2022point}
Alex Nichol, Heewoo Jun, Prafulla Dhariwal, Pamela Mishkin, and Mark Chen.
\newblock Point-e: A system for generating 3d point clouds from complex prompts.
\newblock \emph{arXiv preprint arXiv:2212.08751}, 2022.

\bibitem[Oquab et~al.(2023)Oquab, Darcet, Moutakanni, Vo, Szafraniec, Khalidov, Fernandez, Haziza, Massa, El-Nouby, et~al.]{oquab2023dinov2}
Maxime Oquab, Timoth{\'e}e Darcet, Th{\'e}o Moutakanni, Huy Vo, Marc Szafraniec, Vasil Khalidov, Pierre Fernandez, Daniel Haziza, Francisco Massa, Alaaeldin El-Nouby, et~al.
\newblock Dinov2: Learning robust visual features without supervision.
\newblock \emph{arXiv preprint arXiv:2304.07193}, 2023.

\bibitem[Poole et~al.(2023)Poole, Jain, Barron, and Mildenhall]{pooledreamfusion}
Ben Poole, Ajay Jain, Jonathan~T Barron, and Ben Mildenhall.
\newblock Dreamfusion: Text-to-3d using 2d diffusion.
\newblock In \emph{The Eleventh International Conference on Learning Representations}, 2023.

\bibitem[Reizenstein et~al.(2021)Reizenstein, Shapovalov, Henzler, Sbordone, Labatut, and Novotny]{reizenstein2021co3d}
Jeremy Reizenstein, Roman Shapovalov, Philipp Henzler, Luca Sbordone, Patrick Labatut, and David Novotny.
\newblock Common objects in 3d: Large-scale learning and evaluation of real-life 3d category reconstruction.
\newblock In \emph{Proceedings of the IEEE/CVF International Conference on Computer Vision}, pages 10901--10911, 2021.

\bibitem[Rombach et~al.(2022)Rombach, Blattmann, Lorenz, Esser, and Ommer]{rombach2022high}
Robin Rombach, Andreas Blattmann, Dominik Lorenz, Patrick Esser, and Bj{\"o}rn Ommer.
\newblock High-resolution image synthesis with latent diffusion models.
\newblock In \emph{Proceedings of the IEEE/CVF conference on computer vision and pattern recognition}, pages 10684--10695, 2022.

\bibitem[Shen et~al.(2021)Shen, Gao, Yin, Liu, and Fidler]{shen2021deep}
Tianchang Shen, Jun Gao, Kangxue Yin, Ming-Yu Liu, and Sanja Fidler.
\newblock Deep marching tetrahedra: a hybrid representation for high-resolution 3d shape synthesis.
\newblock \emph{Advances in Neural Information Processing Systems}, 34:\penalty0 6087--6101, 2021.

\bibitem[Shi et~al.(2023{\natexlab{a}})Shi, Chen, Zhang, Liu, Xu, Wei, Chen, Zeng, and Su]{shi2023zero123plus}
Ruoxi Shi, Hansheng Chen, Zhuoyang Zhang, Minghua Liu, Chao Xu, Xinyue Wei, Linghao Chen, Chong Zeng, and Hao Su.
\newblock Zero123++: a single image to consistent multi-view diffusion base model, 2023{\natexlab{a}}.

\bibitem[Shi et~al.(2023{\natexlab{b}})Shi, Wang, Ye, Long, Li, and Yang]{shi2023mvdream}
Yichun Shi, Peng Wang, Jianglong Ye, Mai Long, Kejie Li, and Xiao Yang.
\newblock Mvdream: Multi-view diffusion for 3d generation.
\newblock \emph{arXiv preprint arXiv:2308.16512}, 2023{\natexlab{b}}.

\bibitem[Shim et~al.(2023)Shim, Kang, and Joo]{shim2023diffusion}
Jaehyeok Shim, Changwoo Kang, and Kyungdon Joo.
\newblock Diffusion-based signed distance fields for 3d shape generation.
\newblock In \emph{Proceedings of the IEEE/CVF Conference on Computer Vision and Pattern Recognition}, pages 20887--20897, 2023.

\bibitem[Shue et~al.(2023)Shue, Chan, Po, Ankner, Wu, and Wetzstein]{shue20233d}
J~Ryan Shue, Eric~Ryan Chan, Ryan Po, Zachary Ankner, Jiajun Wu, and Gordon Wetzstein.
\newblock 3d neural field generation using triplane diffusion.
\newblock In \emph{Proceedings of the IEEE/CVF Conference on Computer Vision and Pattern Recognition}, pages 20875--20886, 2023.

\bibitem[Song et~al.(2021)Song, Meng, and Ermon]{songdenoising}
Jiaming Song, Chenlin Meng, and Stefano Ermon.
\newblock Denoising diffusion implicit models.
\newblock In \emph{International Conference on Learning Representations}, 2021.

\bibitem[Sun et~al.(2018)Sun, Wu, Zhang, Zhang, Zhang, Xue, Tenenbaum, and Freeman]{sun2018pix3d}
Xingyuan Sun, Jiajun Wu, Xiuming Zhang, Zhoutong Zhang, Chengkai Zhang, Tianfan Xue, Joshua~B Tenenbaum, and William~T Freeman.
\newblock Pix3d: Dataset and methods for single-image 3d shape modeling.
\newblock In \emph{Proceedings of the IEEE conference on computer vision and pattern recognition}, pages 2974--2983, 2018.

\bibitem[Sun et~al.(2023)Sun, Fang, Wu, Zhang, Zang, Kong, Xiong, Lin, and Wang]{sun2023alphaclip}
Zeyi Sun, Ye Fang, Tong Wu, Pan Zhang, Yuhang Zang, Shu Kong, Yuanjun Xiong, Dahua Lin, and Jiaqi Wang.
\newblock Alpha-clip: A clip model focusing on wherever you want, 2023.

\bibitem[Tang et~al.(2023)Tang, Wang, Zhang, Zhang, Yi, Ma, and Chen]{tang2023make}
Junshu Tang, Tengfei Wang, Bo Zhang, Ting Zhang, Ran Yi, Lizhuang Ma, and Dong Chen.
\newblock Make-it-3d: High-fidelity 3d creation from a single image with diffusion prior.
\newblock In \emph{Proceedings of the IEEE/CVF international conference on computer vision}, pages 22819--22829, 2023.

\bibitem[Tang et~al.(2024)Tang, Chen, Chen, Wang, Zeng, and Liu]{tang2024lgm}
Jiaxiang Tang, Zhaoxi Chen, Xiaokang Chen, Tengfei Wang, Gang Zeng, and Ziwei Liu.
\newblock Lgm: Large multi-view gaussian model for high-resolution 3d content creation.
\newblock \emph{arXiv preprint arXiv:2402.05054}, 2024.

\bibitem[Tochilkin et~al.(2024)Tochilkin, Pankratz, Liu, Huang, , Letts, Li, Liang, Laforte, Jampani, and Cao]{TripoSR2024}
Dmitry Tochilkin, David Pankratz, Zexiang Liu, Zixuan Huang, , Adam Letts, Yangguang Li, Ding Liang, Christian Laforte, Varun Jampani, and Yan-Pei Cao.
\newblock Triposr: Fast 3d object reconstruction from a single image.
\newblock \emph{arXiv preprint arXiv:2403.02151}, 2024.

\bibitem[Valsesia et~al.(2018)Valsesia, Fracastoro, and Magli]{valsesia2018learning}
Diego Valsesia, Giulia Fracastoro, and Enrico Magli.
\newblock Learning localized generative models for 3d point clouds via graph convolution.
\newblock In \emph{International conference on learning representations}, 2018.

\bibitem[Veach(1997)]{VeachPhD}
Eric Veach.
\newblock \emph{{Robust Monte Carlo Methods for Light Transport Simulation}}.
\newblock PhD thesis, Stanford University, 1997.

\bibitem[Walter et~al.(2007)Walter, Marschner, Li, and Torrance]{Walter2007}
Bruce Walter, Stephen~R. Marschner, Hongsong Li, and Kenneth~E. Torrance.
\newblock Microfacet models for refraction through rough surfaces.
\newblock \emph{Eurographics Symposium on Rendering}, 2007.

\bibitem[Wang et~al.(2023)Wang, Du, Li, Yeh, and Shakhnarovich]{wang2023score}
Haochen Wang, Xiaodan Du, Jiahao Li, Raymond~A Yeh, and Greg Shakhnarovich.
\newblock Score jacobian chaining: Lifting pretrained 2d diffusion models for 3d generation.
\newblock In \emph{Proceedings of the IEEE/CVF Conference on Computer Vision and Pattern Recognition}, pages 12619--12629, 2023.

\bibitem[Wang et~al.(2018)Wang, Zhang, Li, Fu, Liu, and Jiang]{wang2018pixel2mesh}
Nanyang Wang, Yinda Zhang, Zhuwen Li, Yanwei Fu, Wei Liu, and Yu-Gang Jiang.
\newblock Pixel2mesh: Generating 3d mesh models from single rgb images.
\newblock In \emph{Proceedings of the European conference on computer vision (ECCV)}, pages 52--67, 2018.

\bibitem[Wang et~al.(2024)Wang, Wang, Chen, Xiang, Chen, Yu, Li, Su, and Zhu]{wang2024crm}
Zhengyi Wang, Yikai Wang, Yifei Chen, Chendong Xiang, Shuo Chen, Dajiang Yu, Chongxuan Li, Hang Su, and Jun Zhu.
\newblock Crm: Single image to 3d textured mesh with convolutional reconstruction model.
\newblock \emph{arXiv preprint arXiv:2403.05034}, 2024.

\bibitem[Wu et~al.(2023{\natexlab{a}})Wu, Johnson, Malik, Feichtenhofer, and Gkioxari]{wu2023multiview}
Chao-Yuan Wu, Justin Johnson, Jitendra Malik, Christoph Feichtenhofer, and Georgia Gkioxari.
\newblock Multiview compressive coding for 3d reconstruction.
\newblock \emph{arXiv preprint arXiv:2301.08247}, 2023{\natexlab{a}}.

\bibitem[Wu et~al.(2017)Wu, Wang, Xue, Sun, Freeman, and Tenenbaum]{wu2017marrnet}
Jiajun Wu, Yifan Wang, Tianfan Xue, Xingyuan Sun, Bill Freeman, and Josh Tenenbaum.
\newblock Marrnet: 3d shape reconstruction via 2.5 d sketches.
\newblock \emph{Advances in neural information processing systems}, 30, 2017.

\bibitem[Wu et~al.(2023{\natexlab{b}})Wu, Zhang, Fu, Wang, Ren, Pan, Wu, Yang, Wang, Qian, et~al.]{wu2023omniobject3d}
Tong Wu, Jiarui Zhang, Xiao Fu, Yuxin Wang, Jiawei Ren, Liang Pan, Wayne Wu, Lei Yang, Jiaqi Wang, Chen Qian, et~al.
\newblock Omniobject3d: Large-vocabulary 3d object dataset for realistic perception, reconstruction and generation.
\newblock \emph{arXiv preprint arXiv:2301.07525}, 2023{\natexlab{b}}.

\bibitem[Wu et~al.(2019)Wu, Wang, Lin, Lischinski, Cohen-Or, and Huang]{wu2019sagnet}
Zhijie Wu, Xiang Wang, Di Lin, Dani Lischinski, Daniel Cohen-Or, and Hui Huang.
\newblock Sagnet: Structure-aware generative network for 3d-shape modeling.
\newblock \emph{ACM Transactions on Graphics (TOG)}, 38\penalty0 (4):\penalty0 1--14, 2019.

\bibitem[Xu et~al.(2024)Xu, Cheng, Gao, Wang, Gao, and Shan]{xu2024instantmesh}
Jiale Xu, Weihao Cheng, Yiming Gao, Xintao Wang, Shenghua Gao, and Ying Shan.
\newblock Instantmesh: Efficient 3d mesh generation from a single image with sparse-view large reconstruction models.
\newblock \emph{arXiv preprint arXiv:2404.07191}, 2024.

\bibitem[Xu et~al.(2019)Xu, Wang, Ceylan, Mech, and Neumann]{xu2019disn}
Qiangeng Xu, Weiyue Wang, Duygu Ceylan, Radomir Mech, and Ulrich Neumann.
\newblock Disn: Deep implicit surface network for high-quality single-view 3d reconstruction.
\newblock \emph{Advances in neural information processing systems}, 32, 2019.

\bibitem[Yang et~al.(2019)Yang, Huang, Hao, Liu, Belongie, and Hariharan]{yang2019pointflow}
Guandao Yang, Xun Huang, Zekun Hao, Ming-Yu Liu, Serge Belongie, and Bharath Hariharan.
\newblock Pointflow: 3d point cloud generation with continuous normalizing flows.
\newblock In \emph{Proceedings of the IEEE/CVF international conference on computer vision}, pages 4541--4550, 2019.

\bibitem[Yariv et~al.(2024)Yariv, Puny, Gafni, and Lipman]{yariv2024mosaic}
Lior Yariv, Omri Puny, Oran Gafni, and Yaron Lipman.
\newblock Mosaic-sdf for 3d generative models.
\newblock In \emph{Proceedings of the IEEE/CVF Conference on Computer Vision and Pattern Recognition}, pages 4630--4639, 2024.

\bibitem[Zhang et~al.(2024)Zhang, Bi, Tan, Xiangli, Zhao, Sunkavalli, and Xu]{zhang2024gs}
Kai Zhang, Sai Bi, Hao Tan, Yuanbo Xiangli, Nanxuan Zhao, Kalyan Sunkavalli, and Zexiang Xu.
\newblock Gs-lrm: Large reconstruction model for 3d gaussian splatting.
\newblock \emph{arXiv preprint arXiv:2404.19702}, 2024.

\bibitem[Zhang et~al.(2018)Zhang, Isola, Efros, Shechtman, and Wang]{zhang2018perceptual}
Richard Zhang, Phillip Isola, Alexei~A Efros, Eli Shechtman, and Oliver Wang.
\newblock The unreasonable effectiveness of deep features as a perceptual metric.
\newblock In \emph{CVPR}, 2018.

\bibitem[Zhao et~al.(2023)Zhao, Liu, Chen, Zeng, Wang, Cheng, FU, Chen, YU, and Gao]{zhao2023michelangelo}
Zibo Zhao, Wen Liu, Xin Chen, Xianfang Zeng, Rui Wang, Pei Cheng, BIN FU, Tao Chen, Gang YU, and Shenghua Gao.
\newblock Michelangelo: Conditional 3d shape generation based on shape-image-text aligned latent representation.
\newblock In \emph{Thirty-seventh Conference on Neural Information Processing Systems}, 2023.

\end{thebibliography}
}
\clearpage
\setcounter{page}{1}
\appendix

\twocolumn[{%
\renewcommand\twocolumn[1][]{#1}%
\begin{center}
    \centering
    \captionsetup{type=figure}
    \includegraphics[width=1.0\textwidth]{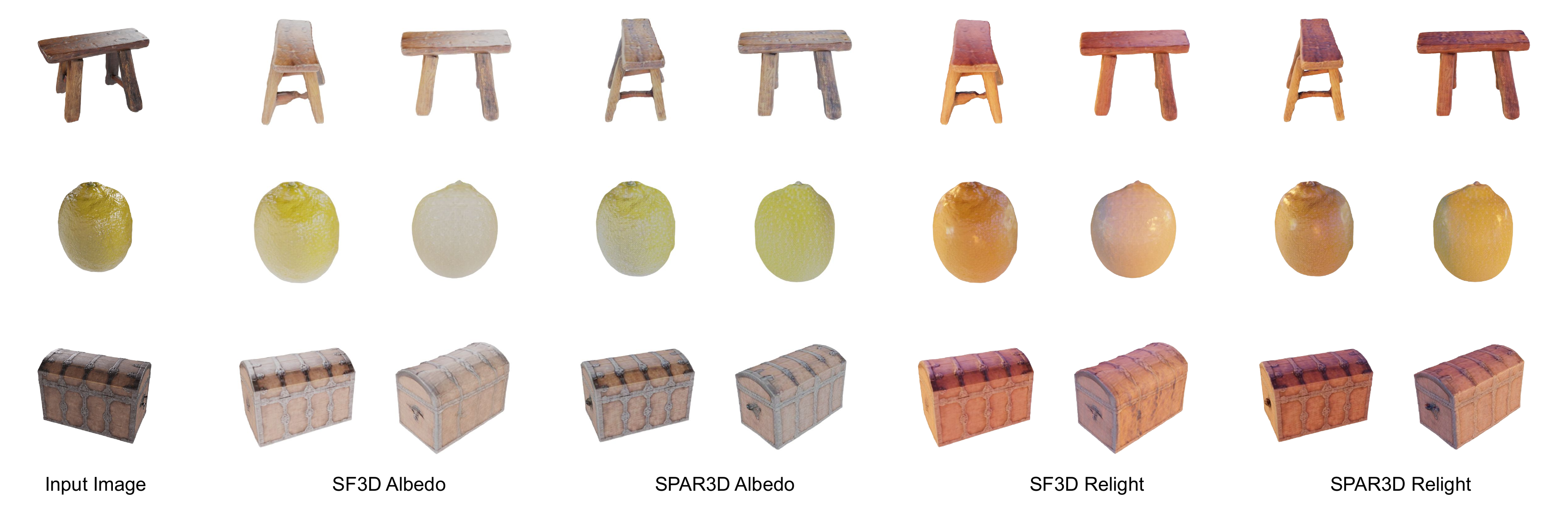}
    \captionof{figure}{\textbf{Decomposition and Relighting Results.} We show decomposed albedo and relighting results of \method in comparison with SF3D. The albedo estimated by \method has less baked-in lighting compared with SF3D and results in better relighting outcomes.}
    \label{fig:supp-decomp}
    \includegraphics[width=1.0\textwidth]{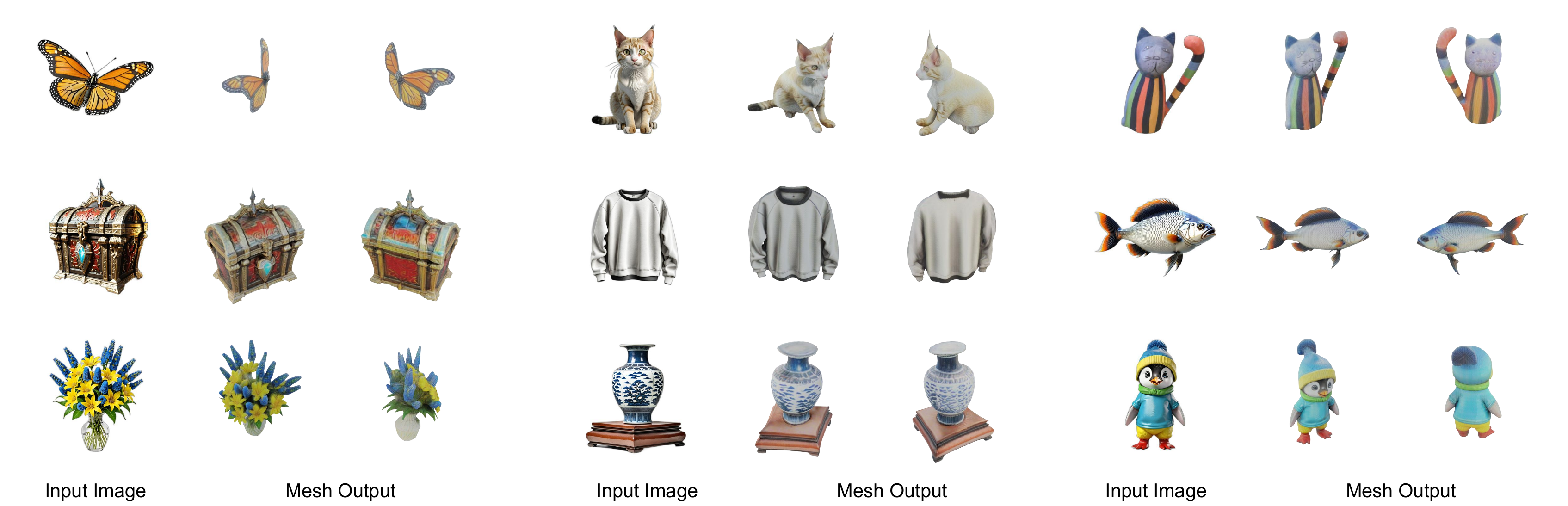}
    \captionof{figure}{\textbf{Additional In-the-wild Results.} We show additional results of \method on in-the-wild images. The reconstructed meshes achieve high fidelity and exhibit great surface details.}
    \vspace{3mm}
    \label{fig:supp-wild}
\end{center}%
}]

This appendix is structured as follows: in~\cref{sec:limitation} we discuss the limitations of our approach; in~\cref{sec:illustration} we provide two additional illustrations of our architecture; in~\cref{sec:decomposition} we show decomposition and relighting results of our model in comparison with SF3D~\cite{sf3d2024}; in~\cref{sec:wild} we present additional in-the-wild results.

\section{Limitations}
\label{sec:limitation}
The main limitations of \method are twofold. First, the point clouds generated during the point sampling stage occasionally exhibit artifacts, such as small surface spikes or detached parts. While these imperfections can typically be remedied through \method's editing capabilities with minimal effort (see Fig. 7 in the main paper), exploring more principled solutions (e.g. improving the denoiser design or diffusion samplers) could further enhance the utility and robustness of our method.

Second, although \method learns material decomposition during training, the accuracy of these decompositions can sometimes be suboptimal. This limitation is primarily due to the inherent ambiguity of inverse rendering from a single image, especially when learned in an unsupervised manner. Unsupervised decomposition learning is useful given the scarcity of 3D assets containing high-quality Physically Based Rendering (PBR) materials and is scalable to real-world multi-view datasets. However, investigating semi-supervised learning techniques may offer a pathway to more plausible material estimations in future work.

\section{Additional Illustrations of our Architecture}
\begin{figure}[h]
\centering
\includegraphics[width=0.48\textwidth]{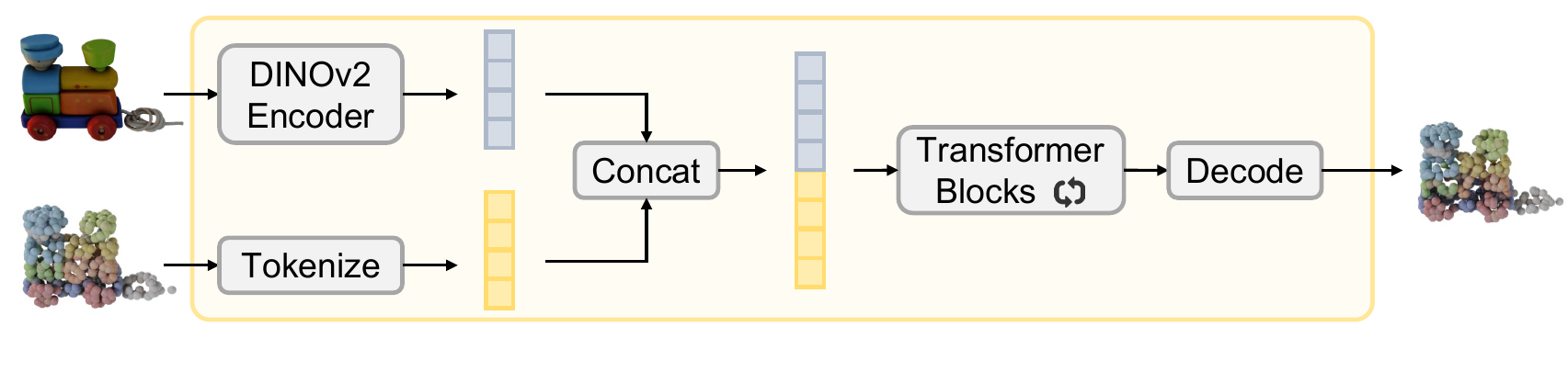}
\caption{\textbf{Point Cloud Denoiser Architecture.} We illustrate the architecture of our point cloud denoiser. The point cloud denoiser takes the noisy point cloud and the image as input, and produces a denoised point cloud. The image and the noisy point cloud are encoded as latent vectors and concatenated together. The concatenated latent vectors are processed by a set of transformer blocks and decoded as the denoised point cloud.}
\label{fig:supp-denoiser}
\end{figure}

\begin{figure}[h]
\centering
\includegraphics[width=0.48\textwidth]{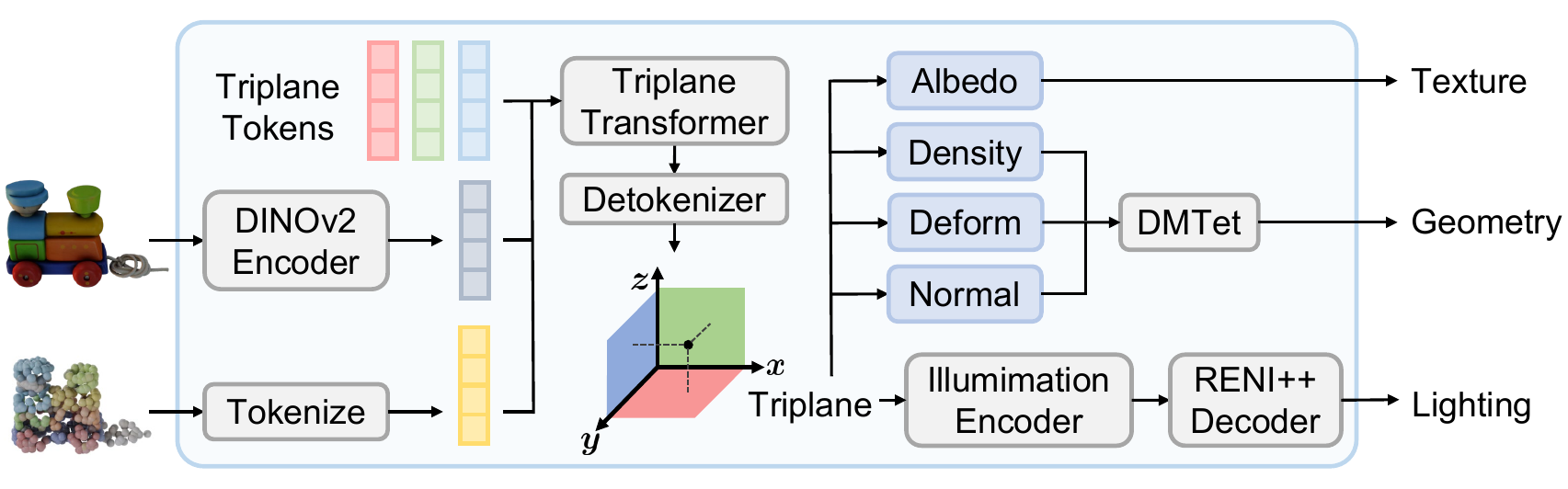}
\caption{\textbf{Meshing Model Architecture.} We illustrate the architecture of our meshing model, which takes the point cloud and the image as input, and produces a textured mesh and an environment map as output. Specifically, the meshing model first encodes the image and the point cloud as latent vectors. The learnable triplane tokens are then processed by the triplane transformer conditioned on the latent vectors. We query the triplane with MLPs to obtain albedo, density, vertex deformation and surface normal, which are converted to a textured mesh using DMTet. The triplane also produces an environment map using the illumination prior from RENI++. The metallic and roughness values are estimated from the image directly and are omitted here for simplicity.}
\label{fig:supp-meshing}
\end{figure}

We show additional illustrations of our point cloud denoiser and our meshing model in~\cref{fig:supp-denoiser} and~\cref{fig:supp-meshing}. We hope these illustrations facilitate a better understanding of our architecture.
\label{sec:illustration}

\section{Decomposition Results}
\label{sec:decomposition}
We show decomposition and relighting results of \method in comparison with SF3D, which is a full regressive method. As shown in~\cref{fig:supp-decomp}, our estimated albedo often has less baked-in lighting artifacts compared with SF3D, which improves the quality of relighting under different illumination conditions. 

\section{Additional In-the-wild Results}
\label{sec:wild}                      
We present additional reconstruction results on in-the-wild images. In~\cref{fig:supp-wild}, we show the reconstructions of \method on images from 3D-Arena (Ebert, 2024). On this data source, \method also achieves high reconstruction quality. This further validates the strong generalization ability of \method.

\end{document}